\documentclass[runningheads]{llncs}

\usepackage{eccv}

\usepackage{eccvabbrv}

\usepackage{graphicx}
\usepackage{booktabs}

\usepackage[accsupp]{axessibility}  % Improves PDF readability for those with disabilities.

\usepackage{hyperref}

\usepackage{orcidlink}
\usepackage{amsmath}
\usepackage{amssymb}
\usepackage{mathtools}
\usepackage{wasysym}
\usepackage{wrapfig} 
\usepackage{soul}
\usepackage[table,xcdraw]{xcolor}
\usepackage{caption}
\usepackage{comment}        % 用于添加长段注释
\usepackage{multirow}
\usepackage{fontawesome5}
\usepackage{tcolorbox}
\begin{document}

\title{MindFlow: Harmonizing Cognitive Semantics and Acoustic Dynamics for Facial Animation Generation in Dyadic Conversations}

\begin{comment}
Old
MindFlow: Decoupling Cognition and Reflexes for Streaming Dyadic Facial Animation

MindFlow: A Chunk-State Paradigm for Streaming Dyadic Facial Animation
MindFlow: Rethinking Dyadic Facial Animation via a Hierarchical Chunk-State Framework
MindFlow: Continuous Emotion and Acoustic Flow for Streaming Conversational Avatars

MindFlow: Harmonizing Cognitive Semantics and Acoustic Dynamics in Dyadic Conversations
MindFlow: Synergizing Emotion States and Acoustic Cues for Streaming Facial Animation
MindFlow: Integrating Deep Semantics and Fine-grained Acoustics for Dyadic Interaction

MindFlow: Seamless Speaking and Listening Animation via Dual-Stream Generative Modeling
MindFlow: Towards Lifelike Dyadic Avatars with Streaming Emotion-Aware Audio-to-Face Generation
MindFlow: Spontaneous and Emotion-Grounded Facial Animation for Dyadic Conversations

\end{comment}

\titlerunning{MindFlow}

\author{
Hejia Chen\inst{1}\thanks{This work was conducted during the internship at Kling Team, Kuaishou Technology
} \and
Haoxian Zhang\inst{2}\thanks{Project lead}\and
Xu He \inst{2} \and
Xiaoqiang Liu \inst{2} \and
Pengfei Wan \inst{2} \and
Shoulong Zhang\inst{3}\textsuperscript{\scriptsize\faEnvelope} \and
Shuai Li\inst{1, 3}\textsuperscript{\scriptsize\faEnvelope}
}

% \orcidlink{0000-1111-2222-3333} 

\authorrunning{H.~Chen et al.}

\institute{Beihang University \and Kling Team, Kuaishou Technology \and Zhongguancun Laboratory \\
\url{https://harryxd2018.github.io/MindFlow/} 
% \and \email{\{abc,lncs\}@uni-heidelberg.de}
}

\maketitle
\begin{figure}
    \centering
    \includegraphics[width=1\linewidth]{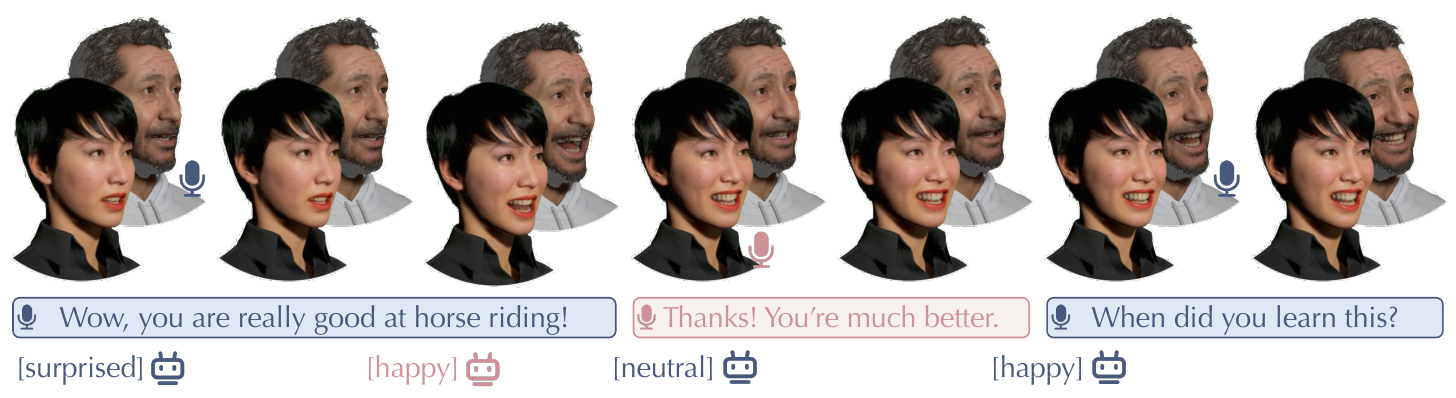}
    \caption{Grounded in the Ventral-Dorsal dual-pathway cognitive model, \textbf{MindFlow} introduces a novel framework for streaming facial animation in dyadic conversations, under which digital avatars simultaneously \textit{perceive} conversational emotions while reflexively \textit{synchronizing} with acoustic rhythms, naturally yielding interactions that are both semantically rich and physically fluid.}
    \label{fig:teaser}
\end{figure}

\begin{abstract}
      Generating lifelike facial animation for dyadic conversations requires reconciling high-level cognitive intent with precise low-level motor reflexes, yet existing methods fall short in the semantic understanding of dialogue context and in precise dynamic control. In this paper, we propose \textbf{MindFlow}, a dual-pathway generative framework inspired by the Ventral-Dorsal pathway model in neuroscience, which decouples generation into two collaborative streams, thereby harmonizing deep semantic reasoning with fine-grained control. In the Ventral module, we transform the conventional Sentence-Action approach into a novel Chunk-State approach that models raw acoustic streams as a context-aware, evolving emotional state chain, capturing subtle paralinguistic nuances and mid-utterance emotional shifts missed by sentence-level modeling. The Dorsal module features a conditional autoregressive flow matching network for high-fidelity facial motion, driven by high-frequency acoustic cues and modulated by emotion states, plus a Selective Acoustic Injector for adaptive audio gating to ensure robustness in talking-and-listening dynamics without interference. Extensive experiments demonstrate that MindFlow achieves superior semantic appropriateness and motion naturalness compared to state-of-the-art baselines.

\end{abstract}

\section{Introduction}
Creating realistic digital avatars capable of fluid, natural conversational interactions is a longstanding fundamental goal within the computer vision and graphics community \cite{ng2024audio2photoreal, peng2025dualtalk, liu2024customlistener}. In particular, generating natural facial dynamics for avatars acting as active conversational interlocutors has garnered significant research attention across multiple fields, including education \cite{prattico2022bot}, entertainment \cite{kyrlitsias2022social}, and social companionship \cite{gao2025effects}. However, despite substantial recent advancements, facial animations of both talking and listening behaviors produced by existing approaches often appear stiff and semantically hollow, due to insufficient semantic understanding of dialogue and limited fine-grained controllability, which limits avatar realism and user engagement.

Compellingly, in neuroscience, a well-established Ventral-Dorsal dual-pathway model~\cite{hickok2007cortical} explains that natural conversation relies on two distinct neural pathways working in parallel: 1) a fast, reflexive \textit{Dorsal pathway} that processes prosodic information and maps acoustic signals directly to articulatory motor areas, and 2) a slow, deliberative \textit{Ventral pathway} that consumes high-level cognitive resources and decodes complex semantic and emotional information. However, existing computational approaches have yet to fully reconcile this functional duality in conversational facial animation generation. Traditional audio-driven methods \cite{ng2022learning, ng2024audio2photoreal, zhu2025infp, peng2025dualtalk} seek to directly map listening behaviors from audio signals, mimicking the Dorsal pathway's functionality, yet they neglect the explicit semantic understanding modeled by the Ventral pathway. This results in animations that are visually reactive but semantically ungrounded and hollow. While recent works have attempted to address this gap by leveraging large language models (LLMs) as proxies for the Ventral pathway to decode sentence-level semantics from the text modality~\cite{liu2024customlistener, zhang2025social}, their sentence-to-action mapping approch suffers from two critical unresolved limitations: 1) \textit{prosodic information loss}: relying solely on text modality inevitably discards critical paralinguistic cues and emotional semantics inherent in the raw conversational audio; and 2) \textit{coarse granularity}: sentence-level segmentation is inherently coarse, which fails to semantically guided fine-grained temporal dynamics of human expression during conversation.

To address the aforementioned limitations, we propose \textbf{MindFlow}, a novel dual-stream framework for high-fidelity conversational facial animation generation in dyadic interactions. Drawing direct inspiration from the Ventral-Dorsal dual-pathway model~\cite{hickok2007cortical} in neuroscience, we decouple conversation-driven facial animation generation into two complementary streams: a cognitive semantic-understanding stream (dubbed {Mind}) and a reflexive flow-matching-based generation stream (dubbed {Flow}), implemented via our proposed Ventral module and Dorsal module, respectively. 

The proposed \textbf{Ventral module} extracts high-level semantic and emotional intent from the evolving dialogue context by incorporating multimodal LLMs (MLLMs). To overcome the problems of coarse granularity and prosodic information loss, rather than using the conventional Sentence-Action approach, we design a novel chunk-state approach that encodes the stream of fixed-window raw dialogue audio chunks and extracts corresponding emotion states. Additionally, we devise a streaming Chain-of-State mechanism that dynamically aggregates historical audio chunks and their corresponding emotion states as an expanding context for the MLLM, enabling temporally consistent reasoning across continuous audio streams. Our Ventral module can preserve delicate paralinguistic cues and encode acoustic context into temporally evolving, continuous emotion states, thereby replacing the static, rigid sentence summaries used in previous work. 

Complementarily, the proposed \textbf{Dorsal module} aims to mimic the sensorimotor reflex system to generate high-fidelity, physically plausible facial motions. To support continuous, variable-length inference and mitigate boundary artifacts \cite{ng2024audio2photoreal, peng2025dualtalk}, we utilize a conditional autoregressive flow matching architecture \cite{yang2025streaming}. However, existing methods often implement a static fusion of dyadic audio signals, diluting effective acoustic features. This diminishes the capability of the audio signal to guide facial motion generation, consequently weakening the avatar's expressiveness during both listening and speaking phases. To explicitly address this critical gap, our key innovation in this module is a novel \textit{Selective Acoustic Injector}. By adaptively filtering high-frequency raw audio features, this component selects the most relevant audio signals to drive facial motions based on the conversational state. Consequently, it guarantees precise, audio-synchronized lip movements during speech and focuses more on the conversational partner's voice during listening phases to generate semantically aligned reactive expressions.

Extensive experiments confirm that our novel dual-stream approach for facial animation generation in the dyadic conversation setting improves lip synchronization and expression accuracy compared with state-of-the-art methods.
In summary, our main contributions are as follows:
\begin{enumerate}
\item We propose {MindFlow}, a novel dual-stream framework inspired by the Ventral-Dorsal pathway model, designed to address the problem of rigid, hollow facial expressions in dyadic conversation animation.
\item We establish a {Chunk-State} approach that enables MLLMs in the Ventral module to analyze the interlocutor's dynamic emotion states from audio streams, effectively preserving audio information and enabling continuous, fine-grained control for expression generation.
\item We design a {Selective Acoustic Injector} in the Dorsal module, which adaptively gates listening and talking dynamics to generate high-quality streaming facial motions seamlessly across both talking and listening states.
\end{enumerate}

\section{Related Work}

\paragraph{Dorsal-Stream-Based Generation. }
Early studies \cite{ng2022learning, liu2024listenformer, siniukov2025ditailistener} and \cite{zhou2022responsive, zhou2025interactive, ng2024audio2photoreal, peng2025dualtalk, guo2025arig, chatziagapi2025av, chen2025midas} build upon talking-head architectures \cite{feng2021learning, he2024co, yang2025streaming, chen2025cafe, zhang2024musetalk, zhong2025anytalker, he2025inpainting, zhao2026emoposeface} by introducing listening branches to extract reactive cues directly from audio signals. These approaches primarily rely on learning low-level correlations between acoustic patterns and facial dynamics. Recent advancements \cite{chen2025dystream, sun2025streamavatar, ki2026avatar} further distill video generation models into autoregressive frameworks to enable online streaming synthesis. While these methods effectively model the temporal causality required for low-latency reflexes, they function similarly to a motor system without a cognitive center. By focusing solely on signal-level mapping, they often fail to capture the long-term semantic coherence and evolving emotion states inherent in dyadic conversations.

\paragraph{Dorsal-Ventral Dual-Stream-Based Generation. }
To incorporate high-level understanding, recent works leverage LLMs to guide facial animation. Some methods \cite{lai2025llm, ng2022learning} fine-tune LLMs \cite{brown2020language} to directly predict reactions, but often suffer from scenario-specific biases due to limited training data. A more prevalent strategy, adopted by \cite{liu2024customlistener, ji2024realtalk, zhang2025social, xue2025echo, wang2025you}, utilizes external LLMs to analyze dialogue content and generate predefined behaviors, such as specific head nods or gaze shifts. We term this the Sentence-Action approach. Although this approach ensures semantic fitness, it suffers from fundamental limitations inherent to its design: relying primarily on the textual modality discards crucial prosodic and emotional cues, while sentence-level reasoning results in coarse-grained, unsynchronized reactions. Similarly, recent holistic conversational frameworks \cite{park2024let, xie2025x, hu2025unitalker, luo2025omniresponse} integrate LLMs to synthesize comprehensive multimodal responses. However, these systems predominantly prioritize the \textit{active talking} phase—generating content and motion during the avatar's turn—often overlooking the continuous, semantically grounded non-verbal feedback required during \textit{active listening}. Unlike these approaches, our framework shifts to a Chunk-State approach, ensuring continuous semantic guidance for both talking and listening roles.

\section{Method}
\subsection{Problem Formulation}
We formulate the task of dyadic facial animation generation from the perspective of an interlocutor $a$ engaged in a symmetric interaction with another interlocutor $b$. Drawing inspiration from the natural operation of the human sensory system, which processes continuous stimuli sequentially, we formulate this task as a causal, streaming generation problem, which naturally diverges from existing non-causal approaches that rely on accessing the full conversational context \cite{ng2024audio2photoreal, peng2025dualtalk, zhu2025infp, liu2024customlistener}. At any time step $t$, the system generates the facial motion $M_a^t$ for interlocutor $a$, conditioning solely on the historical information available up to $t$. The generation process $G$ is modeled as:
\begin{equation}
M_a^t = G(A_a^{\le t}, A_b^{\le t}, S_a^{< t}) ,
\end{equation}
where  $A_a^{\le t}$ and $A_b^{\le t}$ denote continuous raw audio streams, and $S_a^{<t}$ represents the evolving emotion state derived from the interaction history. By adhering to this streaming causality, our formulation mirrors the continuous information flow of biological communication, where future contexts remain inherently inaccessible. Under this biomimetic constraint, the primary challenge is to simultaneously ensure instantaneous audio-visual alignment for dynamic conversational behavior while maintaining the long-term coherence of emotion states.

\subsection{Framework: Ventral-Dorsal Dual Pathway}

\begin{figure}[htbp]
    \centering
    \includegraphics[width=1\linewidth]{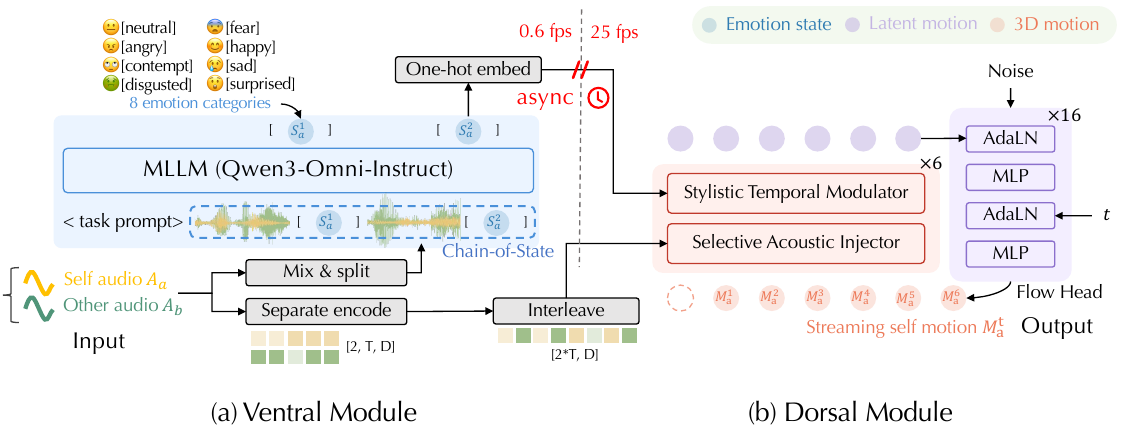}
    \caption{Inspired by the \textcolor[HTML]{3488BB}{Ventral}-\textcolor[HTML]{EF835F}{Dorsal} dual-pathway model, \textbf{MindFlow} generates lifelike conversational facial animations (both listening and speaking) driven by continuous dialogue audio \textcolor[HTML]{EBCD8A}{$A_a$} and \textcolor[HTML]{A9C693}{$A_b$}. 1) The \textcolor[HTML]{3488BB}{Ventral module}  functions as a cognitive semantic perceiver, leveraging an MLLM with a streaming \textit{Chain-of-State} to decode evolving emotion states chunk-by-chunk, conditioned on the historical cognitive trajectory. 2) The \textcolor[HTML]{EF835F}{Dorsal module} functions as a reflexive sensory-motor executor. It incorporates a \textit{Selective Acoustic Injector} that uses motion queries to adaptively gate unmixed audio streams, and a \textit{Stylistic Temporal Modulator} that injects the Ventral emotion states as semantic guidance. These combined features condition an autoregressive flow-matching backbone for the continuous generation of high-fidelity, synchronous facial motions.}
    \label{fig:pipeline}
\end{figure}

Inspired by the biological Ventral-Dorsal dual-pathway model, which separately processes semantic comprehension and sensory-motor reflexes, we construct the MindFlow framework (Fig. \ref{fig:pipeline}). Specifically, MindFlow incorporates a Ventral module and a Dorsal module to emulate these respective functions, where the two modules collaborate seamlessly to ensure the avatar's facial dynamics are both contextually appropriate and consistently engaging, effectively overcoming the issue of hollow expressions common in traditional audio-driven methods.

To bridge these two modules, we introduce a novel \textit{Chunk-State approach}. Previous methods \cite{liu2024customlistener, zhang2025social} typically adopt a \textit{Sentence-Action approach}, which performs behavior reasoning based on sentence-level textual content. To plan proper movements, they often force the reasoning model to output verbose motion descriptions burdened with complex spatial and temporal details, leading to temporally mismatched and coarse-grained responses. In contrast, our approach operates directly on raw audio streams at a much finer chunk-level granularity. Because it reasons at the scale of short chunks, our Ventral module naturally achieves more precise alignment without needing to explicitly predict temporal details. Furthermore, aligning with the biological dual-pathway model, we argue that intricate spatial expression details should not be pre-planned cognitively. Instead, the Ventral module acts solely as the cognitive foundation, continuously ingesting audio chunks to decode and update a fluid emotion state. Concurrently, the Dorsal module functions as the sensory-motor executor. Conditioned on the Ventral module's semantic guidance and instantaneous acoustic cues, it takes over the generation of fine spatial details, translating them into high-fidelity, synchronous reflexes including accurate lip articulation, rhythmic head movements, and dynamic expressions. Formally, this dual-pathway process at time step $t$ can be formulated as:
\begin{equation}
\begin{cases}
S_a^k = \text{Ventral}(A_a^{\le k}, A_b^{\le k}, S_a^{< k}) \\
M_a^t = \text{Dorsal}(A_a^{\le t}, A_b^{\le t}, M_a^{< t}, S_a^{\lfloor t/w \rfloor})
\end{cases}, 
\end{equation}
where $k = \lfloor t/w \rfloor$ denotes the discrete chunk index, and $w$ represents the chunk window size. Through this chunk-state collaboration, MindFlow bypasses the stiffness of sentence-level textual planning, guaranteeing that the avatar not only comprehends the semantic flow to react reasonably but also remains physically "alive" and highly responsive at every moment.

\subsection{Ventral Module: Streaming Cognitive State Modeling}
The Ventral module functions as the cognitive anchor of the MindFlow framework, emulating the biological Ventral pathway responsible for semantic and emotional comprehension. Its primary objective is to maintain an emotion state derived from the evolving conversation history. By operating on a coarse temporal scale, it distills the continuous audio streams into context-aware cognitive states via semantic reasoning. 

Unlike the Sentence-Action approach \cite{liu2024customlistener, zhang2025social} that relies on sentence-level reasoning, our Chunk-State approach analyzes the interlocutor's emotional variations using fine-grained audio chunks as the fundamental unit. Operating directly on the audio modality offers two critical advantages. First, it avoids the inevitable loss of prosodic information that occurs when converting speech to textual transcripts. Second, the smaller temporal window of audio chunks significantly enhances the temporal granularity of the semantic guidance provided by the Ventral module. Consequently, we turn to MLLMs \cite{Qwen3-Omni, funaudiochat2025} to perform direct semantic and emotional analysis natively within the audio modality.

\begin{figure}[htbp]
    \centering
    \includegraphics[width=0.9\linewidth]{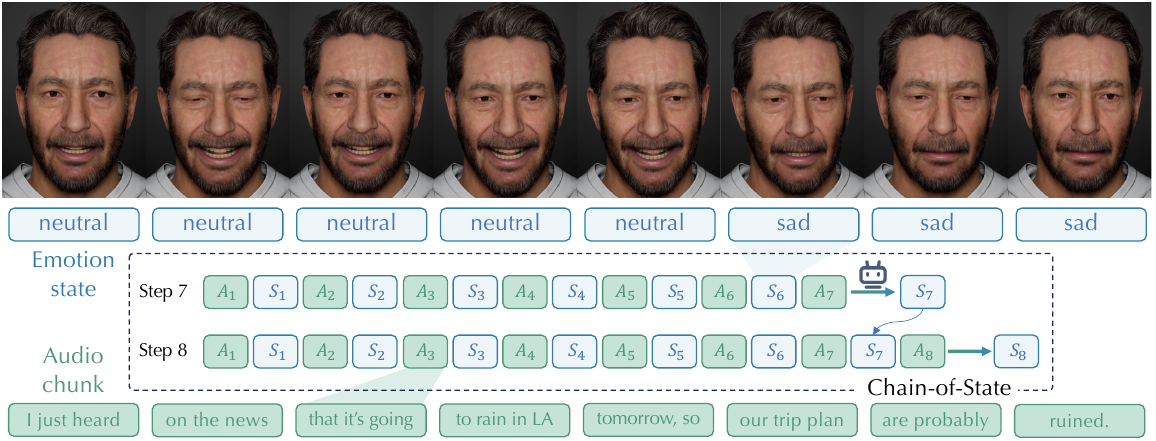}
    \caption{Visualization of Ventral module reasoning and corresponding generated result. The MetaHuman plugin in Unreal Engine is utilized for rendering.}
    \label{fig:result}
\end{figure}

However, existing MLLMs are predominantly tailored for turn-based question-answering tasks, posing a significant challenge for streaming continuous conversations. If we were to independently query the MLLM to analyze the emotion of each individual chunk, the rich conversational context would be completely destroyed. Conversely, if we simply concatenated the current chunk with historical chunks as the input query, the MLLM would lack explicit awareness of its past emotion states, frequently leading to unstable and temporally inconsistent predictions for the current chunk. To resolve this dilemma, we propose a streaming Chain-of-State mechanism (Fig. \ref{fig:result}) that explicitly models the evolving emotion state while simultaneously preserving the complete history of the interaction. 

Formally, at step $k$, the Ventral module predicts the current state $S_a^k$ conditioned not only on the current audio chunk $A_k$ (as mixed chunk $A_a^{wk:wk+w}$ and $ A_b^{wk:wk+w}$ from both interlocutors) but also on the history of past audio inputs and previously inferred states $(A_a^{\le wk}, A_b^{\le wk}, S_a^{< k})$. By progressively appending the predicted emotion state to the context alongside the audio stream, this formulation establishes an expanding memory of both the sensory input and the cognitive trajectory. These derived emotion states are concurrently transmitted to the Dorsal module to guide facial motion generation. Ultimately, this Chain-of-State mechanism effectively mitigates the inherent instability of memory-less MLLM inferences; by explicitly conditioning on the trajectory of past states, it preserves contextual continuity, ensuring the avatar's emotional expressions evolve smoothly, robustly, and coherently throughout the continuous interaction.

\subsection{Dorsal Module: Reflexive Motion Generation}
Acting as the sensory-motor executor of our framework, the Dorsal module emulates the biological Dorsal pathway, tasked with generating immediate subconscious reflexes driven by high-frequency acoustic cues. 
To achieve high-fidelity motion that is both responsive and physically coherent, we design a unified architecture comprising an autoregressive Transformer backbone tailored with mechanism-specific injectors, and a flow matching head for generating expressive and diverse facial motion \cite{cen2025ready, yang2025streaming}.

\paragraph{Stylistic Temporal Modulator.} To ensure that the generated reflexes adhere to the emotion state obtained by the Ventral module, we introduce the Stylistic Temporal Modulator. Implemented as a specialized Transformer encoder layer, this module enforces semantic consistency while preserving temporal continuity. The emotion state $S_a^k$ from the Ventral module is first embedded and appended to the input hidden motion sequence at step $t=wk$. We employ a masked causal attention mechanism where the emotion state embedding is visible to subsequent hidden motion $H_m^{wk:wk+w}$ as guidance, while preventing information leakage from future frames. Guided by this semantic-aware emotion state, the Dorsal module generates rich high-frequency dynamics, ensuring that the avatar's expressions during the conversation are both vivid and engaging.

\paragraph{Selective Acoustic Injector.} Following the sensory-motor integration function of the biological Dorsal pathway, the Dorsal module takes the audio streams of both interlocutors ($A_a$ and $A_b$) as input. Audio-driven methods \cite{ng2024audio2photoreal, peng2025dualtalk, zhu2025infp} commonly adopt concatenation-based fusion to early-mix these distinct signals. Specifically, they concatenate the audio features and map them through an MLP prior to cross-attention injection, which can be formulated as:
\begin{equation}
    F_{\text{concat}} = \text{Attn}(H_m, A_{\text{fuse}}, A_{\text{fuse}}), \quad \text{where } A_{\text{fuse}} = \text{MLP}(\text{Concat}(A_a, A_b)).
\end{equation}
This early-mix strategy prematurely entangles disparate acoustic sources into a shared latent space. Since the attention mechanism is forced to query from this mixed context, it obscures the distinct functional boundaries required for talking articulation versus listening reaction, inevitably causing signal dilution.

To address this, we introduce the Selective Acoustic Injector, implemented as a modified Transformer decoder layer. Instead of early mixing, the audio features from both interlocutors are temporally interleaved to form a composite, source-independent acoustic context $A_{\text{ctx}} = \text{Interleave}(A_a, A_b)$. We formulate the injection process as a query-response mechanism where the motion history $H_m$ actively attends to relevant acoustic cues:
\begin{equation}
    F_{\text{inject}} = \text{Attn}(H_m, A_{\text{ctx}}, A_{\text{ctx}}) = \text{Softmax}\left(\frac{Q_m K_{\text{ctx}}^\top}{\sqrt{d}}\right) V_{\text{ctx}}.
\end{equation}

\begin{figure}[htbp]
    \centering
    \includegraphics[width=0.75\linewidth]{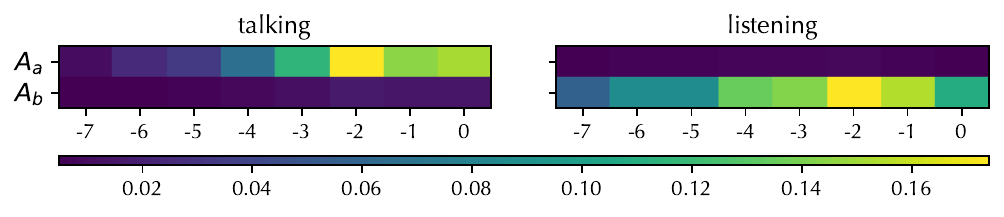}
    \caption{Attention map showing motion query focusing on relevant audio.}
    \label{fig:attn}
\end{figure}

By exposing the unmixed audio streams directly to the attention mechanism, this design allows the injector to learn a dynamic gating policy. Crucially, we do not impose any explicit supervision or speaker labels to direct this attention mechanism. Instead, the dynamic gating strategy acts as an emergent behavior optimized purely by the flow matching objective. As visualized in the attention heatmap (Fig. \ref{fig:attn}), the model exhibits a clear, unsupervised phase transition: while generating the motion of interlocutor $a$, the query predominantly locks onto the \textit{self} track ($A_a$) for precise lip-sync during active speech, but spontaneously shifts focus to the \textit{other} track ($A_b$) to trigger reactive behaviors during listening. This confirms that our injector naturally internalizes the turn-taking dynamics of dyadic conversation, dynamically filtering the acoustic context based solely on motion causality.

\paragraph{Autoregressive Transformer Backbone.} To effectively integrate the stylistic and acoustic representations, our backbone is constructed by alternately stacking the aforementioned modules. Specifically, the network comprises $L = 6$ identical blocks, where each block sequentially applies the Stylistic Temporal Modulator and the Selective Acoustic Injector. During training, the backbone operates under a teacher-forcing strategy, utilizing the historical ground-truth motion sequence $M_a^{<t}$ to predict the motion condition $C_a^t$ for the current frame. Formally, we define the end-to-end conditioning generation process as:
\begin{equation}
C_a^t = \text{Backbone}\left(M_{<t}, S_a^t, A_{ctx}\right),
\end{equation}
where the function $\text{Backbone}(\cdot)$ encapsulates the feature embedding, $L$ alternating blocks, and a final linear projection. The output $C_a^t$ provides the precise expression context required for the subsequent flow matching process.

\paragraph{Flow Matching Head.} To capture the inherent expressive diversity of realistic interactions, we discard deterministic regression in favor of a generative flow matching approach \cite{lipman2022flow}. We formulate facial motion generation as learning a transport map from a standard Gaussian distribution $\pi_0 = \mathcal{N}(0, I)$ to the data distribution $\pi_1$. The flow matching head, parameterized as an MLP $v_\theta$, learns to predict the velocity field that drives the ordinary differential equation:
\begin{equation}
    dZ_\tau = v_\theta(Z_\tau, \tau| C) d\tau,
\end{equation}
where $Z_\tau$ is the motion state at timestep $\tau \in [0, 1]$, and $C$ represents the conditioning context from the transformer backbone. During training, we optimize the flow matching objective:
\begin{equation}
    \mathcal{L}_{flow} = \mathbb{E}_{\tau \sim \mathcal{U}[0,1], Z_0 \sim \pi_0, Z_1 \sim \pi_1} [\| v_\theta(Z_\tau, \tau | C) - (Z_1 - Z_0) \|^2]. 
\end{equation}

Unlike diffusion models \cite{ho2020denoising} that require iterative denoising, flow matching learns straight-line trajectories, allowing us to approximate the solution using a simple 5-step Euler solver during inference. This efficiency is critical, enabling our Dorsal module to synthesize high-fidelity, non-deterministic facial details with a rapid responsiveness characteristic of the biological Dorsal pathway.

\section{Experiments}
\subsection{Implementation Details}
The Dorsal module is trained on a combination of public datasets totaling approximately 20 hours: HDTF~\cite{zhang2021flow} provides fundamental motion priors; VICOX\footnote{\url{https://project.mhzhou.com/vico}} strengthens listening behavior and interaction modeling; and MEAD~\cite{MEAD} and VICO~\cite{zhou2022responsive} capture emotional talking and listening behaviors, respectively. We adopt a two-stage strategy: the model is first pre-trained on HDTF and VICOX for 90k steps to establish basic talking and conversational behaviors, then fine-tuned on MEAD and VICO for 30k steps to improve emotion awareness and expressiveness ($\sim 4$ days in total). Both stages use the Adam optimizer with a batch size of 64, a peak learning rate of $1e^{-5}$ with a cosine schedule and a $1\%$ warmup, and no weight decay.  The audio encoder~\cite{baevski2020wav2vec} is frozen throughout both stages. Facial movements are encoded as 51-dimensional ARKit blendshape coefficients for expressions and 3D Euler angles for head pose, extracted via MediaPipe \cite{lugaresi2019mediapipe} and FSA-Net \cite{yang2019fsa}, respectively.

During inference, the two modules run asynchronously. The Ventral module processes each 1.5 s audio chunk in $1.38\pm0.10\mathrm{s}$, while the Dorsal module generates facial reactions at 25 FPS in real time, reusing $S_a^k$ until the next state $S_a^{k+1}$ arrives. The full system requires 59 GB VRAM and sustains real-time output over 2-minute sequences without memory growth.

\subsection{Comparison with Prior Methods}
\subsubsection{Quantitative Evaluation}
\paragraph{Metrics. } 
We introduce a comprehensive set of evaluation metrics to systematically assess the quality of generated facial movements in both talking and listening states. To evaluate motion realism, we employ the Fréchet Distance (FD). For assessing lip synchronization during talking phases, we utilize SyncNet scores (SyncC and SyncD), while Mean Squared Error (MSE) is adopted to measure expression accuracy within the listening context. Notably, we adapt SyncNet \cite{chung2017out, li2024latentsync} from video-based synchronization evaluation to the 3D modality.

\begin{table}[htbp]
\centering
\setlength{\tabcolsep}{2.5pt} 
\renewcommand{\arraystretch}{1.1} 

\caption{\textbf{Quantitative comparison with SOTA methods.} We evaluate performance on two distinct interaction modes: (a) talking generation and (b) listening responsiveness. \colorbox[HTML]{C6E3D8}{Green} and \colorbox[HTML]{E8F4F0}{Cyan} highlights denote the best and second-best results.}
\label{tab:sota_comparison}

% ------ 左边子表: Speak Evaluation ------
\begin{subtable}[b]{0.48\linewidth}
    \centering
    \caption{Evaluation on talking state.}
    \label{tab:talk-eval}
    % \scriptsize % 如果表格还是太宽，可以取消这一行的注释
    \begin{tabular}{@{}lcccc@{}}
    \toprule
                                  & \multicolumn{2}{c}{SyncNet}                                   & \multicolumn{2}{c}{FD $\downarrow$}                          \\ \cmidrule(lr){2-3} \cmidrule(l){4-5} 
    \multirow{-2}{*}{Method}      & {\scriptsize SyncD} $\downarrow$ & {\scriptsize SyncC} $\uparrow$ & {\scriptsize Exp}             & {\scriptsize Pose}           \\ \midrule
    EmoTalk                       & 0.429                            & 0.412                          & 23.52                         & -                            \\
    UniTalker                     & 0.480                            & 0.300                          & 29.88                         & -                            \\ \midrule
    DualTalk                      & 0.467                            & 0.346                          & 26.06                         & 0.18                         \\
    A2P                           & \cellcolor[HTML]{E8F4F0}0.341& \cellcolor[HTML]{E8F4F0}0.519& \cellcolor[HTML]{E8F4F0}17.64 & \cellcolor[HTML]{E8F4F0}0.03 \\
    Ours                          & \cellcolor[HTML]{C6E3D8}0.333& \cellcolor[HTML]{C6E3D8}0.520& \cellcolor[HTML]{C6E3D8}15.76 & \cellcolor[HTML]{C6E3D8}0.01 \\ \bottomrule
    \end{tabular}
\end{subtable}
\hfill % 撑开间距
% ------ 右边子表: Listen Evaluation ------
\begin{subtable}[b]{0.48\linewidth}
    \centering
    \caption{Evaluation on listening state.}
    \label{tab:listen_eval}
    \begin{tabular}{@{}lcccc@{}}
    \toprule
             & \multicolumn{2}{c}{FD$\downarrow$} & \multicolumn{2}{c}{MSE$\downarrow$} \\ \cmidrule(lr){2-3} \cmidrule(l){4-5} 
    Method   & {\scriptsize Exp} & {\scriptsize Pose} & {\scriptsize Exp} & {\scriptsize Pose} \\ \midrule
    L2L      & 33.93             & 0.06               & 0.93              & \cellcolor[HTML]{C6E3D8}0.01 \\
    RLHG     & 39.02             & 0.07               & 0.86              & \cellcolor[HTML]{C6E3D8}0.01 \\
    DIM      & 23.88             & 0.06               & 0.70              & \cellcolor[HTML]{C6E3D8}0.01 \\
    DualTalk & 22.27             & 0.05               & 0.58              & \cellcolor[HTML]{C6E3D8}0.01 \\
    A2P      & \cellcolor[HTML]{E8F4F0}14.24 & \cellcolor[HTML]{E8F4F0}0.03 & \cellcolor[HTML]{E8F4F0}0.34 & \cellcolor[HTML]{C6E3D8}0.01 \\
    Ours     & \cellcolor[HTML]{C6E3D8}13.86 & \cellcolor[HTML]{C6E3D8}0.03 & \cellcolor[HTML]{C6E3D8}0.30 & \cellcolor[HTML]{C6E3D8}0.01 \\ \bottomrule
    \end{tabular}
\end{subtable}
\end{table}

\paragraph{Talking Performance. } We compare the talking performance of our method with previous approaches on the HDTF testset \cite{zhang2021flow}. Specifically, we compare with talking head models EmoTalk and UniTalker \cite{peng2023emotalk, fan2024unitalker}, which use the same motion representation as ours. Meanwhile, we re-train the dyadic methods \cite{peng2025dualtalk, ng2024audio2photoreal} to align with our motion representation. As shown in Tab. \ref{tab:talk-eval}, MindFlow achieves state-of-the-art performance across all metrics, encompassing lip synchronization, facial expression fidelity, and head movement realism. Notably, our Dorsal module operates via autoregressive rollout to generate motion in a frame-by-frame streaming manner, supporting continuous inference for long conversational audio and overcoming the fixed-length limitations ($\sim 10s$) of previous methods.

\paragraph{Listening Performance. } We follow the evaluation protocol of DualTalk to assess the quality of generated listening behaviors on the VICO testset. As shown in Tab. \ref{tab:listen_eval}, our model achieves lower scores in terms of FD and MSE, indicating that it generates more accurate and appropriate facial movements during the listening state. We attribute this improvement to: 1) the introduction of semantic guidance as a prior for motion generation, making the generated results more appropriate; and 2) the use of a generative network architecture, which is capable of producing more diverse facial motion patterns and effectively mitigates the over-smoothing issue commonly observed in discriminative models.

\begin{figure}[t]
    \centering
    \includegraphics[width=1\linewidth]{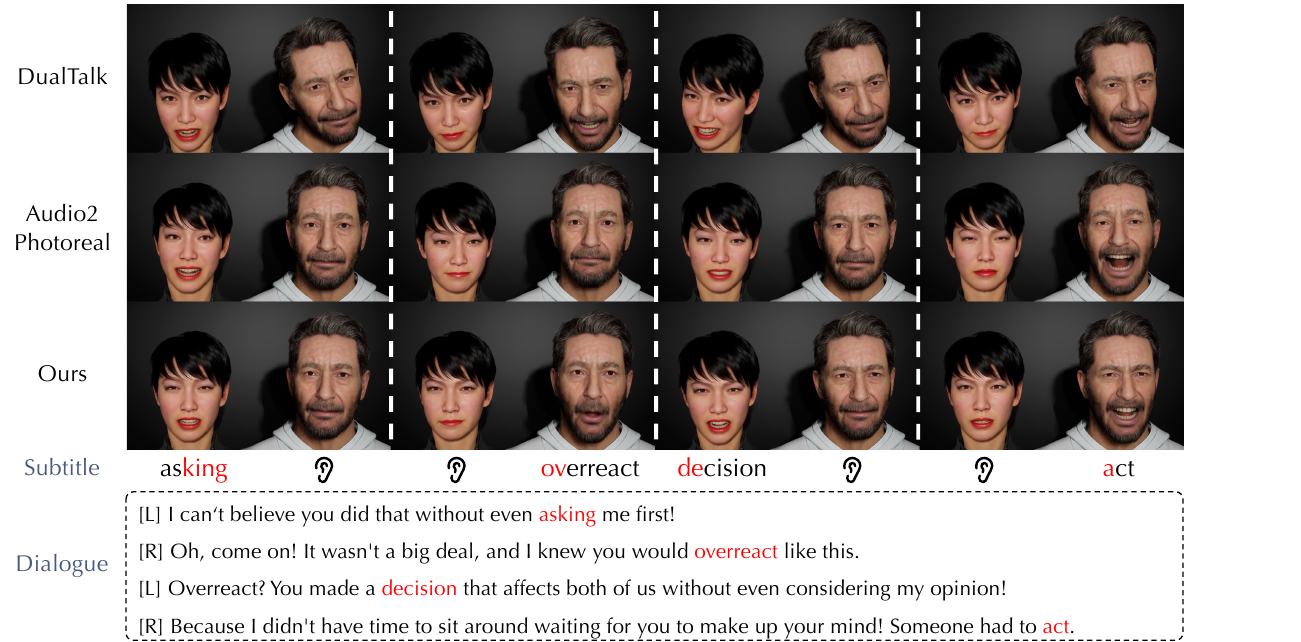}
    \caption{\textbf{Performance under conversational tension}: MindFlow captures the intense atmosphere to exhibit semantically and temporally appropriate expressions, whereas baseline methods suffer from unstable and hollow expressions due to the boundary artifacts and a lack of semantic guidance.}
    \label{fig:compare}
\end{figure}

\subsubsection{Qualitative Evaluation}
We further conduct a qualitative comparison to assess the proposed method from the perspectives of motion naturalness and lip-sync consistency. First, we adopt DualTalk and A2P as baseline Dorsal-only methods for comparison. Fig. \ref{fig:compare} presents representative results generated by different methods over multiple dialogue turns, including alternating talking and listening states. Compared to DualTalk, whose head pose exhibits noticeable discrepancies in motion amplitude and frequency between talking and listening states, our method produces smoother and more natural head movements across different interaction modes. In contrast to A2P, which struggles to maintain stable facial expressions over extended durations, our approach is able to preserve controllable and coherent emotional expressions over long sequences. We attribute this limitation of A2P to its fixed-length diffusion transformer structure, which hinders the long-term consistency of facial motions via sliding-window inference, whereas the autoregressive flow matching generation structure in the Dorsal module enables more stable and reliable inference for long-duration facial animation.

\begin{figure}[h]
    \centering
    \includegraphics[width=1\linewidth]{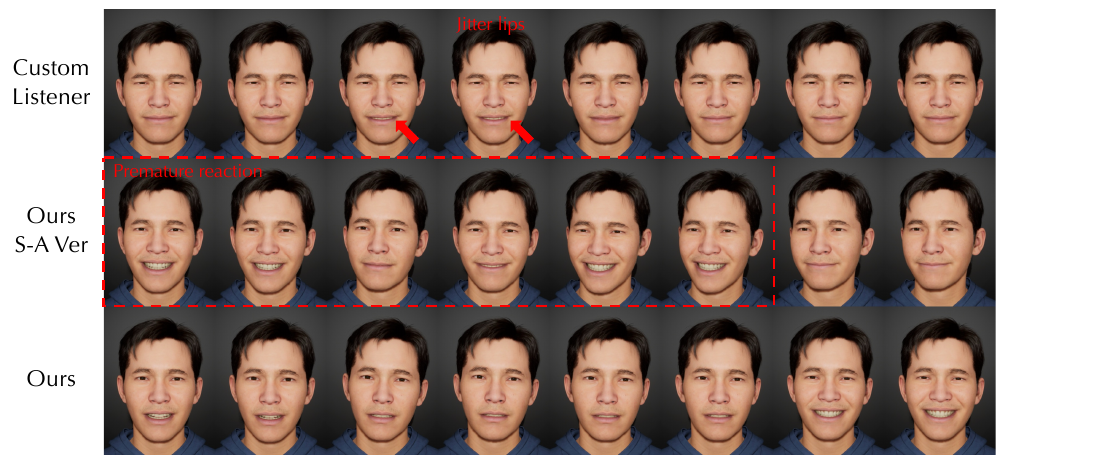}
    \caption{Compared to the Sentence-Action approach, our proposed Chunk-State approach yields reactions with more appropriate timing and style.}
    \label{fig:vs_CL_SA}
\end{figure}

Second, we compare with CustomListener\footnote{Obtained from the CustomListener homepage.}, which leverages LLM for semantic understanding on sentence-level textual dialogue. As illustrated in Fig. \ref{fig:vs_CL_SA}, our approach generates appropriate, well-timed reactions (\ie, a smile at the end). In contrast, CustomListener fails to produce noticeable reactions and exhibits static facial expressions, a limitation stemming from its coarse sentence-level action constraints. 
This indicates that our Chunk-State approach possesses a distinct advantage in temporal granularity over the Sentence-Action approach.

\begin{figure}[htbp]
    \centering
    \includegraphics[width=\linewidth]{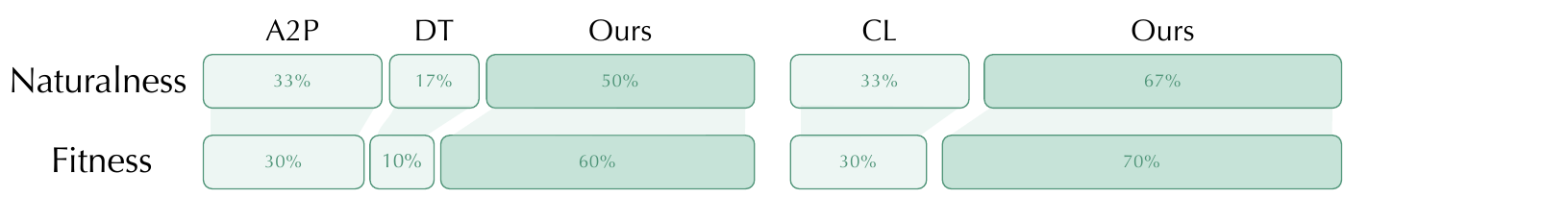}
    \caption{Results of the perception study.}
    \label{fig:user_study}
\end{figure}

Finally, we conduct a perception study for a more comprehensive evaluation. We designed a survey utilizing a comparative format, where evaluators were asked to select the best generation results based on two criteria: the naturalness and fitness of the generated movements. 
To ensure a blind evaluation and prevent bias, the correspondence between the methods and the presented options was randomized. 
As illustrated in Fig. \ref{fig:user_study}, out of the 24 collected evaluations, our model consistently outperforms previous methods in both naturalness and fitness.

\subsection{Ablation Study}
\paragraph{Semantic Guidance. } 

We analyze the impact of the Ventral module's semantic guidance provided to the Dorsal module through quantitative ablation experiments. Specifically, on the VICO testset, we provide the Dorsal module with three distinct types of guidance signals: a random state, a fixed state (sentence-level GT), and the emotion state derived from the Ventral module. As shown in Tab. \ref{tab:ablation_slow}, compared to random or fixed guidance, the evolving emotion state extracted from the dialogue by the Ventral module effectively guides the Dorsal module to generate more contextually appropriate facial expressions. This indicates the guidance from the Ventral module is temporally dynamic, fine-grained, and aligned with the ground-truth facial expressions.

\paragraph{Chunk-State Approach. } 
We compare our method against a Sentence-Action variant of our own model, which degrades the Chain-of-State mechanism to sentence-level reasoning. As shown in Fig. \ref{fig:vs_CL_SA}, although the variant can also yield appropriate emotional states, the resulting reactions are temporally unsynchronized and premature. This distinctly underscores the critical role of fine-grained semantic reasoning, 
which is further validated in Tab. \ref{tab:ablation_slow} (\textit{v.s.} S-A).

\begin{table}[htbp]
\centering
\caption{Results of component-wise ablation study.} % 这里的总标题
\label{tab:overall_ablation}
\setlength{\tabcolsep}{4pt} 

% 左侧第一个子表
\begin{subtable}[c]{0.52\linewidth}
    \centering
    \caption{Ventral module.} % 子标题 (a)
    \label{tab:ablation_slow}
    \begin{tabular}{lcccc}
    \toprule
                             & \multicolumn{2}{c}{FD$\downarrow$}                           & \multicolumn{2}{c}{MSE $\downarrow$}                         \\ \cline{2-5} 
    \multirow{-2}{*}{Method} & Exp                           & Pose                         & Exp                          & Pose                         \\ \midrule
    Random                   & 15.21                         & 0.03                         & 0.36                         & 0.01                         \\
    Fixed                    & \cellcolor[HTML]{E8F4F0}14.15 & \cellcolor[HTML]{E8F4F0}0.03 & \cellcolor[HTML]{E8F4F0}0.32 & \cellcolor[HTML]{E8F4F0}0.01 \\ \hline
    S-A                      & 14.39                         & 0.03                         & 0.33                   & 0.01                               \\ \hline
    Ours               & \cellcolor[HTML]{C6E3D8}13.86 & \cellcolor[HTML]{C6E3D8}0.03 & \cellcolor[HTML]{C6E3D8}0.30 & \cellcolor[HTML]{C6E3D8}0.01 \\ \bottomrule
    \end{tabular}
\end{subtable}
\hfill % 自动拉开间距
% 右侧第二个子表
\begin{subtable}[c]{0.45\linewidth}
    \centering
    \caption{Selective Acoustic Injector (SAI).} % 子标题 (b)
    \label{tab:sai-ablation}
    \renewcommand{\arraystretch}{1.05}
    \begin{tabular}{lcc@{\hspace{4pt}}c}
    \toprule
    Method                & SAI & SyncD$\downarrow$             & SyncC$\uparrow$               \\ \midrule
                          & w/o & 0.341                         & 0.519                         \\
    \multirow{-2}{*}{A2P} & w/  & \cellcolor[HTML]{C6E3D8}0.331 & \cellcolor[HTML]{C6E3D8}0.526 \\ \hline
                          & w/o & 0.350                         & 0.424                         \\
    \multirow{-2}{*}{Ours}  & w/  & \cellcolor[HTML]{E8F4F0}0.333 & \cellcolor[HTML]{E8F4F0}0.520 \\ \bottomrule
    \end{tabular}
\end{subtable}
\end{table}

\begin{figure}[htbp]
    \centering
    \includegraphics[width=0.75\linewidth]{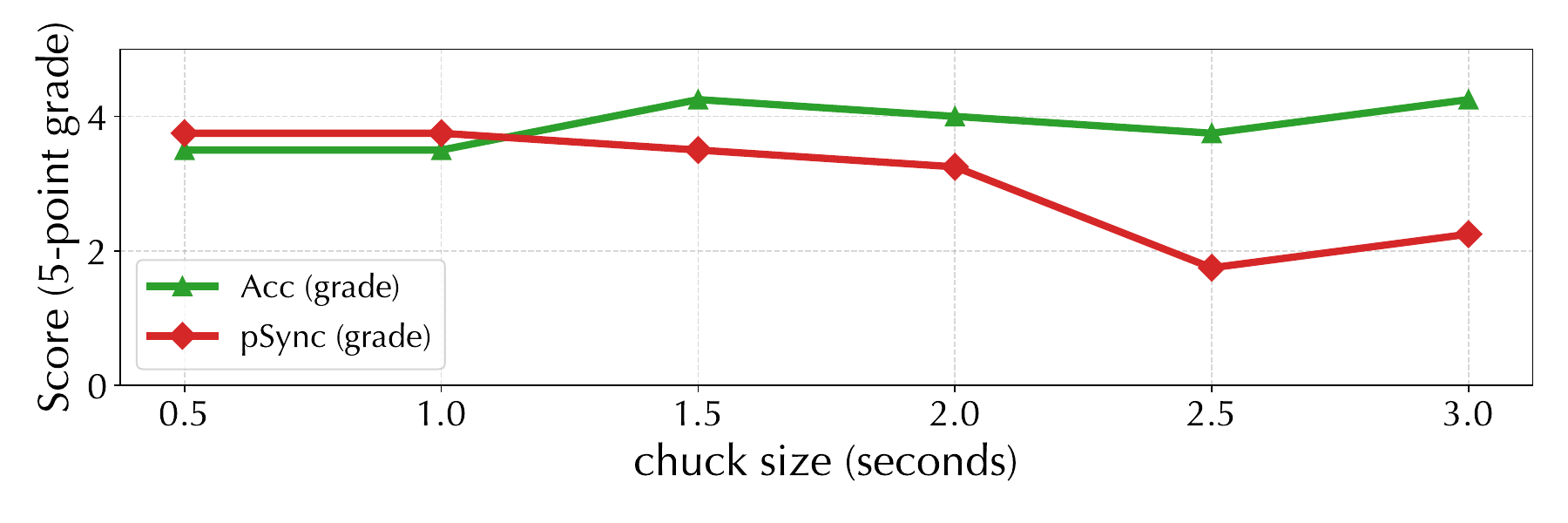}
    \caption{Impact of chunk size.}
    \label{fig:ablation_LLM}
\end{figure}

\paragraph{Reason Unit Length. } The window size of the chunk $w$ serves as a critical hyperparameter governing the trade-off between the granularity of emotion state and stability. To identify the optimal setting, we conduct a user study to assess emotion prediction accuracy (Acc) and perceptual synchronicity (pSync) under different chunk sizes. As shown in Fig. \ref{fig:ablation_LLM}, while increasing the chunk size improves accuracy, which is likely due to the incorporation of richer temporal context, it degrades the perceptual synchronicity. To balance these trade-offs, we empirically set the chunk size to 1.5 seconds.

\paragraph{Selective Acoustic Injector.}
We validate this injector by comparing our method and a retrained A2P \cite{ng2024audio2photoreal} against two cross-ablated variants: (1) A2P augmented with our proposed injector, and (2) our method using the A2P acoustic module. As shown in Tab. \ref{tab:sai-ablation}, incorporating the injector consistently improves lip-sync quality across architectures. This demonstrates its effectiveness in selectively modeling cross-temporal acoustic features and providing more informative guidance.
Moreover, the superior performance of the A2P-augmented variant over our causal method aligns with the known advantage of bidirectional approaches, further validating the consistent effectiveness of the proposed injector.

\section{Conclusion}
We presented MindFlow, a novel dual-stream framework that bridges the gap between high-level cognitive reasoning and precise motor reflexes for streaming dyadic facial animation. By introducing a hierarchical Chunk-State approach, MindFlow overcomes the limitations of rigid sentence-level constraints. Ventral module leverages a streaming Chain-of-State mechanism to extract continuous, fine-grained emotional semantics directly from audio chunks. Concurrently, the Dorsal module employs a Selective Acoustic Injector and a flow matching head to generate high-fidelity, synchronous facial dynamics seamlessly across both talking and listening phases. Extensive experiments demonstrate that MindFlow significantly outperforms state-of-the-art methods in semantic appropriateness and motion naturalness, enabling more lifelike and engaging conversational avatars.

\section{Limitations and Future Work}
While MindFlow significantly advances audio-driven dyadic facial animation, it presents certain limitations that pave the way for future research. Primarily, our current framework relies exclusively on auditory inputs to infer the interlocutor's emotional and semantic states. In natural face-to-face interactions, visual information, \eg, the interlocutor's eye contact, facial expressions, and body language, is equally crucial for conveying non-verbal intent and emotional nuances. Relying solely on the acoustic modality may occasionally limit the avatar's capacity to react to silent yet semantically rich behaviors. Therefore, a key direction for future work is to extend our framework to accommodate multimodal sensory inputs. By seamlessly integrating the interlocutor's visual cues alongside the raw audio streams, the framework can achieve a more holistic contextual understanding, enabling the generation of even more empathetic, accurate, and contextually appropriate facial responses.

% \clearpage  % TODO FINAL: This \clearpage needs to be removed from both review and camera-ready versions.

% \section*{Acknowledgements}
% Please insert your acknowledgments here.

\section*{Acknowledgments}
This work was supported by the National Key R\&D Program of China (2023YF F1203803), Zhongguancun Laboratory, and the National Natural Science Foundation of China (62502469, 62525204).

\appendix
\section*{Supplementary Material}
\section{Overview}
% 在这份补充材料中我们提供更多技术细节，在Section \ref{sec:details}中介绍和模型实现有关的Details，在Section \ref{sec:exp}中介绍实验的更多细节，我们同时提供了视频以支持本文工作。
In this supplementary material, we provide additional technical details. Section \ref{sec:details} elaborates on the implementation of our proposed model, while Section \ref{sec:exp} presents further experimental results and configurations. Furthermore, we include a supplementary video\footnote{\url{https://www.youtube.com/watch?v=tQfgv38swJs}} to provide visual demonstrations of our work.

\section{Implementation Details}
\label{sec:details}
\subsection{Ventral Module}
We demonstrate the prompt design that leverages Multimodal Large Language Models (MLLMs) within the Ventral module to implement a Chain-of-State mechanism.

\begin{tcolorbox}[fontupper=\scriptsize]
You are a streaming audio conversational emotion analyzer designed for online, chunk-level inference, serving an expressive conversational digital human system.

You will receive multiple audio chunks from the same conversation in chronological order.
The conversation always and exclusively involves two speakers: one male and one female.

\textbf{Core Task}\\
After receiving each new audio chunk, you must:\\
- Determine the dominant emotional state of the male speaker at the current moment (i.e., at the end of the current audio chunk), based on all historical audio content received up to this point.\\
- Even when the male speaker is silent and in a listening state, you must infer his emotional reaction based on the female speaker’s speech content, rather than defaulting to [neutral].\\

Emotion Label Set (you must strictly select one) \\
- [angry]\\
- [contempt]\\
- [disgusted]\\
- [fear]\\
- [happy]\\
- [sad]\\
- [surprised]\\
- [neutral]\\

\textbf{Decision Rules} (critical constraints affecting expressiveness)\\
- You must jointly consider both speech content (semantics) and vocal prosody (intonation, tone, stress, rhythm, energy variation, etc.) when determining emotion.\\
- When the male speaker is listening, prioritize analyzing the female speaker’s speech content and tone, and infer the emotional impact on the male speaker as his internal emotional response.\\
- Whenever any perceivable emotional cue exists (whether from the male speaker’s own speech or emotional stimulation caused by the female speaker’s dialogue), you should select the closest specific emotion label instead of [neutral].\\
- [neutral] should be used only when: 1) The overall conversation remains stable and lacks semantic or prosodic cues that may trigger emotional change in the male speaker; 2) The current audio chunk contains insufficient emotion-related information to determine an emotional tendency.\\
- When a mild but recognizable emotional tendency is present, prefer selecting a low-intensity specific emotion label rather than reverting to [neutral].\\

\textbf{Continuity and Emotional Evolution Constraints} \\
- You can and must reference your previous emotion predictions from earlier audio chunks.\\
- Emotion should be treated as a continuously evolving state rather than independent classification results.\\
- If the current audio chunk continues the emotional trend from the previous moment, you should maintain or smoothly transition to a similar emotion instead of frequently reverting to [neutral].\\
- Emotional changes should be gradual and explainable, with significant transitions occurring only when clear semantic or prosodic changes are observed.\\

\textbf{Output Format }(must be strictly followed) \\
- For each audio chunk, output exactly one emotion label.\\
- Do not output any explanation, reasoning process, punctuation, or additional text.
\end{tcolorbox}

\subsection{Dorsal Module}
To mitigate the exposure bias introduced by teacher-forcing and the resulting error accumulation, and to enable stable inference over arbitrarily long audio-driven dialogues, we carefully design both the training and inference procedures.

\begin{figure}[htbp]
    \centering
    \includegraphics[width=1\linewidth]{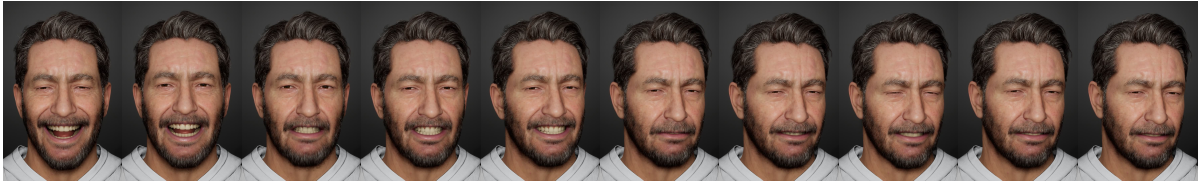}
    \caption{\textbf{A failure case of head-pose drift}: the drift appears after only a few inference steps. Frames are displayed at $2.5$ fps for clarity.}
    \label{fig:icml-pose-drift}
\end{figure}

\paragraph{Head-pose representation.} We observe that when the Dorsal module is trained to directly predict head poses represented by 3D Euler angles, noticeable drift emerges during inference, as the predicted orientation gradually rotates toward an extreme angle (as shown in Fig. \ref{fig:icml-pose-drift}). We attribute this phenomenon to the module becoming overly dependent on historical motion during training, thereby encouraging it to perpetuate the previous head-pose trajectory at inference time.

To address this issue, we train the Dorsal module to predict the angular velocity of the head pose instead of the absolute Euler angles. This formulation not only prevents the module from exploiting such shortcuts but also aligns better with the intrinsic correlation between audio and head-pose dynamics. We further find that angular velocity exhibits stronger coupling with the speech signal, leading to more coherent and speech-consistent head movements during talking. During inference, for all frames except the first, the current head pose is obtained by adding the predicted angular velocity to the head pose of the previous frame. For the first frame, the module still directly predicts the absolute head pose rather than its angular velocity, which keeps the training and inference procedures consistent.

\paragraph{Historical perturbation. }
To further prevent the Transformer backbone from over-relying on historical motion and learning undesirable shortcuts, we follow the inspiration of Chen \etal \cite{chen2024diffusion} and add Gaussian noise $N$ to the input $M$ as in Eq. \ref{eq: noised}. During training, $\sigma$ is randomly sampled from $[0.01, 0.05]$ to ensure that the noise does not disrupt the overall motion structure. During inference, $\sigma$ is fixed to $0.02$.
\begin{equation}
    M^{\text{noised}} = (1-\sigma) M + \sigma N
    \label{eq: noised}
\end{equation}

This perturbation encourages a balanced reliance between historical motion and audio features when generating the current frame's motion.

\paragraph{Talking dynamic expressiveness.} We aim to enhance the expressiveness of the generated motions by increasing the influence of audio features. This is achieved by adjusting the cross-modal injection coefficient $\lambda_{cross}$ within the Selective Acoustic Injector. In the cross-attention module where audio conditions are injected (Eq.~\ref{eq: cross_lambda}), we introduce an additional scaling factor before the multi-head attention residual term:
\begin{equation}
    \label{eq: cross_lambda}
    H_m^{\text{out}} =
    H_m^{\text{in}} +
    \lambda_{cross} \cdot
    \mathrm{Softmax}\!\left(
    \frac{(\mathrm{Norm}(H_m))(A_{ctx})^\top}{\sqrt{d}}
    \right) A_{ctx},
\end{equation}
where $H_m$ and $A_{ctx}$ are intermediate motion latent and audio feature, respectively. By increasing $\lambda_{cross}$, the model relies more heavily on audio features when autoregressively predicting the next-frame motion, thereby strengthening the influence of acoustic cues on expression generation. In practice, $\lambda_{cross}$ is set to $2.5$.

\section{Experiments}
\label{sec:exp}
\paragraph{SyncNet Metrics. } Building upon LatentSync \cite{li2024latentsync}, we adapt the SyncNet model to our motion representation to evaluate lip-sync accuracy. Specifically, we replace the original visual branch with a series of MLP layers to accommodate the new modality. This modified network is then trained on a high-quality 3D mocap talking dataset. The resulting model calculates evaluation metrics following the original SyncNet methodology \cite{chung2017out}.

\paragraph{Talking Performance}
We further report the evaluation results on the HDTF test set. Using the LVE and Diversity metrics \cite{FaceFormer, guo2020action2motion} , we compare the proposed method with previous works from two perspectives: lip-sync accuracy and expression diversity. As shown in Tab. \ref{tab:lips-eval}, MindFlow achieves state-of-the-art performance in both lip accuracy and expression diversity on the benchmark. We attribute this improvement to the autoregressive flow-matching-based architecture adopted in the Dorsol module.

\begin{table}[htbp]
\centering
\caption{Further evaluation on the HDTF testset. }
\label{tab:lips-eval}
\begin{tabular}{@{}lccccc@{}}
\toprule
Method                                  & EmoTalk                       & UniTalker & Audio2photoreal& DualTalk & Ours                           \\ \midrule
LVE$\downarrow$                         & \cellcolor[HTML]{E8F4F0}6.212 & 7.866     & 9.676                          & 8.806    & \cellcolor[HTML]{C6E3D8}6.305  \\
Diversity $\uparrow$ ($\times 10^{-2}$) & 5.558                         & 2.242     & \cellcolor[HTML]{E8F4F0}21.332 & 2.074    & \cellcolor[HTML]{C6E3D8}24.197 \\ \bottomrule
\end{tabular}
\end{table}

\begin{figure}[htbp]
    \centering
    \includegraphics[width=0.5\linewidth]{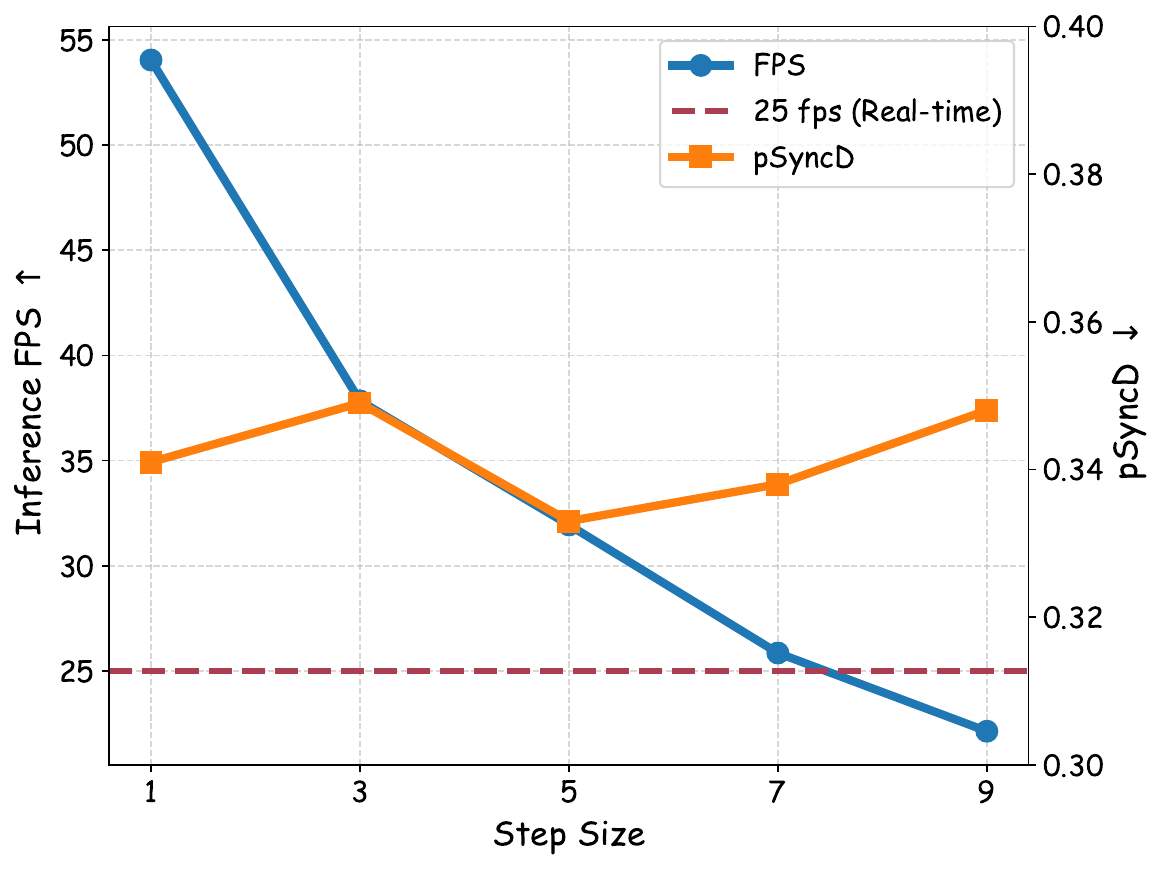}
    \caption{Impact of sampling step on synthesis quality and inference latency}
    \label{fig:ablation_step}
\end{figure}

\paragraph{Impact of Sampling Steps.} To determine the optimal trade-off of the Dorsal module between synthesis quality and inference latency, we analyze the effect of the number of sampling steps on the lip-synchronization performance. We evaluate the results using SyncD and inference speed (FPS). As shown in Fig. \ref{fig:ablation_step}, the relationship between sampling steps and SyncD is non-monotonic. At low step counts ($\leq 5$), the discretization error is dominant, leading to under-convergence of the ODE solver and coarse lip shapes. Increasing the steps initially improves alignment; however, we observe a performance degradation when sampling steps exceed $5$. This counterintuitive phenomenon can be attributed to the accumulation of prediction errors in the vector field over a longer integration trajectory, which introduces temporal jitter and high-frequency artifacts that negatively impact the SyncNet scores. Consequently, we adopt 5-step sampling as the optimal setting, achieving the lowest SyncD score while maintaining a real-time throughput.

\section{Ethics Statement}
MindFlow introduces a dual-pathway framework for generating streaming 3D facial animations in dyadic conversations. By harmonizing cognitive semantics with acoustic dynamics, this model enables digital avatars to achieve natural, context-aware, and expressive interactions. While this technology offers significant benefits in domains such as social companionship, virtual education, and accessibility tools, it also presents potential ethical risks. The high-fidelity 3D facial parameter generation and coefficient-driven techniques could be misused to create unauthorized digital avatar interactions, fabricated character performances, or identity impersonation. To mitigate these risks, the application of this technology must be grounded in informed consent and respect for individual privacy. We advocate for transparent disclosure or data provenance technologies to ensure users can distinguish AI-generated digital content. Strict adherence to ethical standards is essential to ensure that conversational animation technology contributes positively to society and enhances the human-computer interaction experience.

\bibliographystyle{splncs04}
\bibliography{main}
\end{document}